\documentclass[10pt,twocolumn,letterpaper]{article}
\usepackage{cvpr}
\usepackage{times}
\usepackage{epsfig}
\usepackage{graphicx}
\usepackage{amsmath}
\usepackage{amssymb}
\usepackage{color}
\newcommand{\argmax}{\mathop{\rm argmax}\limits} 
\usepackage[pagebackref=true,breaklinks=true,letterpaper=true,colorlinks,bookmarks=false]{hyperref}

\cvprfinalcopy 

\def\cvprPaperID{2149} 

\ifcvprfinal\fi
\begin{document}

\title{Mining Discriminative Triplets of Patches for Fine-Grained Classification}

\author{Yaming Wang$^{\dag}$, \; Jonghyun Choi$^{\ddag}$, \; Vlad I. Morariu$^{\dag}$, \; Larry S. Davis$^{\dag}$\\
$^{\dag}$University of Maryland, College Park, MD 20742, USA\\
$^{\ddag}$Comcast Labs DC, Washington, DC 20005, USA\\
{\tt\small \{wym, jhchoi, morariu, lsd\}@umiacs.umd.edu}
}

\maketitle
\thispagestyle{empty}

\begin{abstract}
Fine-grained classification involves distinguishing between similar sub-categories based on subtle differences in highly
localized regions; therefore, accurate localization of discriminative regions remains a major challenge. We
describe a patch-based framework to address this problem. We introduce triplets of patches with geometric
constraints to improve the accuracy of patch localization, and automatically mine discriminative
geometrically-constrained triplets for classification. The resulting approach only requires object
bounding boxes. Its effectiveness is demonstrated using four publicly available fine-grained datasets, on which it  
outperforms or achieves comparable performance to the state-of-the-art in classification. 
\end{abstract}

\section{Introduction}\label{sec1}
The task of fine-grained classification is to recognize sub-ordinate categories belonging to the same super-ordinate
category \cite{cub2010, cub2011, dog1, car196}. The major challenge is that fine-grained objects share similar overall appearance and
only have subtle differences in highly localized regions. To effectively and accurately find these discriminative regions, some
previous approaches utilize humans-in-the-loop \cite{human2, bb, human1}, or require semantic part annotations
\cite{dog2, berg1, berg2, berg3, ningzhang1, ningzhang2, ningzhang3} or 3D models \cite{yenliang, car196}.
These methods are effective, but they require extra keypoint/part/3D annotations from humans, which are often expensive to obtain. On the other
hand, recent research on discriminative mid-level visual elements mining
\cite{patch1, patch2, patch4, patch3, patch5} automatically finds discriminative patches or regions from a huge
pool and uses the responses of
those discriminative elements as a mid-level representation for classification. However, this approach has mainly been applied
to scene classification and not typically to fine-grained classification. This is probably due to the
fact that the discriminative patches needed for fine-grained categories need to be more accurately localized than for scene
classification.

To avoid the cost of extra
annotations, we propose a {\it patch-based} approach for fine-grained classification that overcomes the
difficulties of previous discriminative mid-level approaches. Our approach requires only object bounding
box annotations and belongs to the category of weakly-annotated fine-grained classification \cite{fgtemplate,
bangpeng, fgalign, fgmultask, symbiotic, har, b_cnn, krause15}.

Two issues need to be addressed. The first issue is accurately localizing discriminative patches without requiring extra
annotations. Localizing a single patch based only on its
 appearance remains challenging due to noisy
backgrounds, ambiguous repetitive patterns and pose variations. Instead, we localize {\it triplets} of patches
with the help of geometric constraints. Previously, triple or higher-order constraints have been used in image matching
and registration \cite{trimatch1, trimatch2, trimatch3}, but they have not been integrated into a patch-based classification
framework. Our triplets consist of three appearance descriptors and two simple, but efficient, geometric constraints,
which can be used to remove accidental detections that would be encountered if patches were localized individually.
One attractive property of triplets over simpler models (\eg, pairs) is that they can be used to model rich geometric
relationships while providing additional invariance (\eg, to rotation) or robustness (\eg, to perspective
changes).
\begin{figure}
\centering
\includegraphics[width=0.49\textwidth]{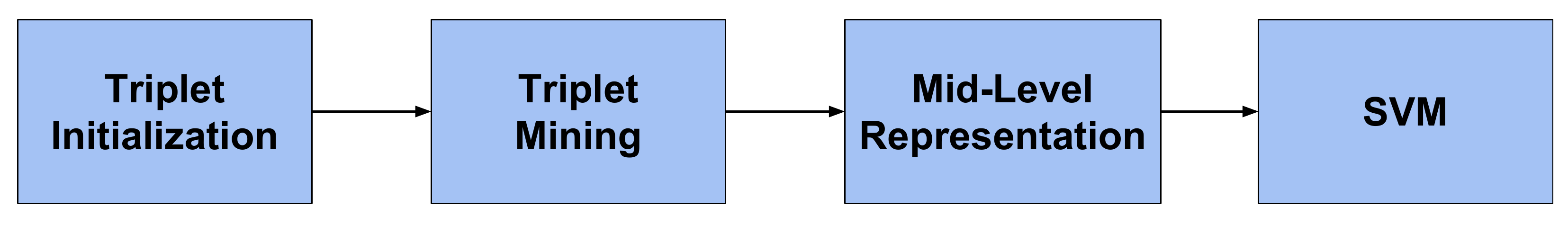}
\caption{\label{fig4} Summary of our discriminative triplet mining framework. Candidate triplets are initialized
from sets of neighboring images and selected by how discriminative they are across the training set. The mid-level
representation consists of the maximum responses of the selected triplets with geometric constraints, which is fed into
a linear SVM for classification.}
\end{figure}

The second issue is automatically discovering discriminative geometrically-constrained triplets from the huge pool of
all possible triplets of patches. The key insight is that fine-grained objects share similar
overall appearance. Therefore, if we retrieve the nearest neighbors of a training image, we obtain samples from different
classes with almost the same pose, from which potentially discriminative regions can be found. Similar ideas have been
adopted in \cite{loclearn, fgalign, ptrans, ellf}, but they aggregate results obtained from local neighborhood
processing without further analysis across the whole dataset. In contrast, our
discriminative triplet mining framework uses sets of overall similar images to propose potential triplets, and 
only selects good ones by measuring their discriminativeness using the whole training set or a large portion of it.

We evaluate our approach on four publicly available fine-grained datasets and obtain comparable results to the state-of-the-art without expensive annotation.

\section{Triplets of Patches with Geometric Constraints} \label{sec2}
In this section, we discuss two geometric constraints encoded within triplets of patches and describe a triplet detector
incorporating these constraints. 

Suppose we have three patch templates $T_A$, $T_B$ and $T_C$, with their centers located
at points $A$, $B$ and $C$, respectively. Each template $T_i$ ($i \in \{A, B, C\}$) can be a feature vector extracted from a single patch or
an averaged feature vector of patches from the corresponding locations of several positive samples. Given an image, let
$A'$, $B'$ and $C'$ be three patches possibly corresponding to $A$, $B$ and $C$, respectively.

\subsection{Order Constraint and Shape Constraint}\label{sec2_1}
\begin{figure}
\centering
\includegraphics[width=0.39\textwidth]{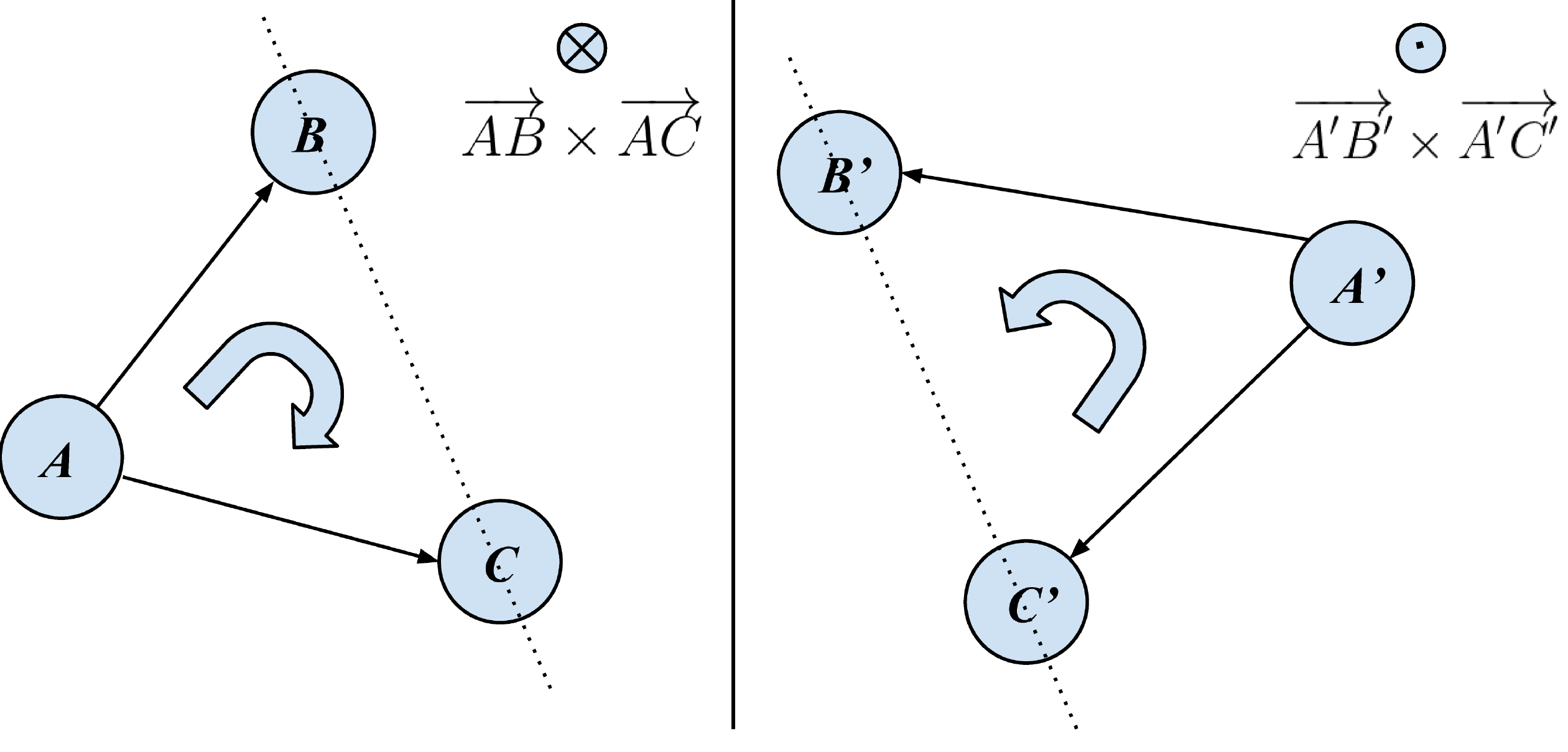}
\caption{\label{fig2} Visualization of the order constraint. Top: Patch $A$, $B$ and $C$ are arranged in clockwise order. The
direction of $\protect\overrightarrow{AB} \times \protect\overrightarrow{AC}$ points into the paper, and $G_{ABC} = 1$. Bottom: $A'$,
$B'$ and $C'$ are arranged in counterclockwise order. The direction of the cross product points out of the paper, and
$G_{A'B'C'} = -1$.}
\end{figure}

{\bf The order constraint} encodes the ordering of the three patches (Figure \ref{fig2}). For triplet $\{A, B,
C\}$, consider the two vectors $\overrightarrow{AB}$ and $\overrightarrow{AC}$. Treating them as three dimensional
vectors with the third dimension being $0$, it follows that
\begin{equation}\label{eq1}
\overrightarrow{AB} \times \overrightarrow{AC} = (0, 0, Z_{ABC}).
\end{equation}
Let  
\begin{equation}\label{eq+2}
G_{ABC} = {\rm sign}(Z_{ABC}),
\end{equation}
which indicates whether the three patches are arranged clockwise
($G_{ABC} = 1$) or counterclockwise ($G_{ABC} = -1$), as can be seen in Figure \ref{fig2}. There is a side-test 
interpretation of this constraint \cite{corresp}. If we fix two patches, say $B$ and $C$, $G_{ABC} = 1$ means that 
$A$ lies on the left side of the line passing between $B$ and $C$, while $G_{ABC} = -1$ indicates $A$ is on the right side.
Therefore, a simple penalty function between $\{A, B, C\}$ and $\{A', B', C'\}$ based on the order constraint can be
defined as
\begin{equation}\label{eq2}
p_o(G_{ABC}, G_{A'B'C'}) = 1 - \eta_o{\rm \bf l}(G_{A'B'C'} \neq G_{ABC}),
\end{equation}
where $0 \le \eta_o \le 1$ controls how important the order constraint is, and the indicator function is defined as
\begin{align}\label{eq3}
{\rm \bf l}(G_{A'B'C'} \neq G_{ABC}) = 
\begin{cases}
1,~ G_{A'B'C'} \neq G_{ABC} \\
0,~ G_{A'B'C'} = G_{ABC}.
\end{cases}
\end{align}
Intuitively, $p_o$ penalizes by $\eta_o$ when $\{A', B', C'\}$ violates the order constraint defined by $\{A, B, C\}$.

\begin{figure}
\centering
\includegraphics[width=0.35\textwidth]{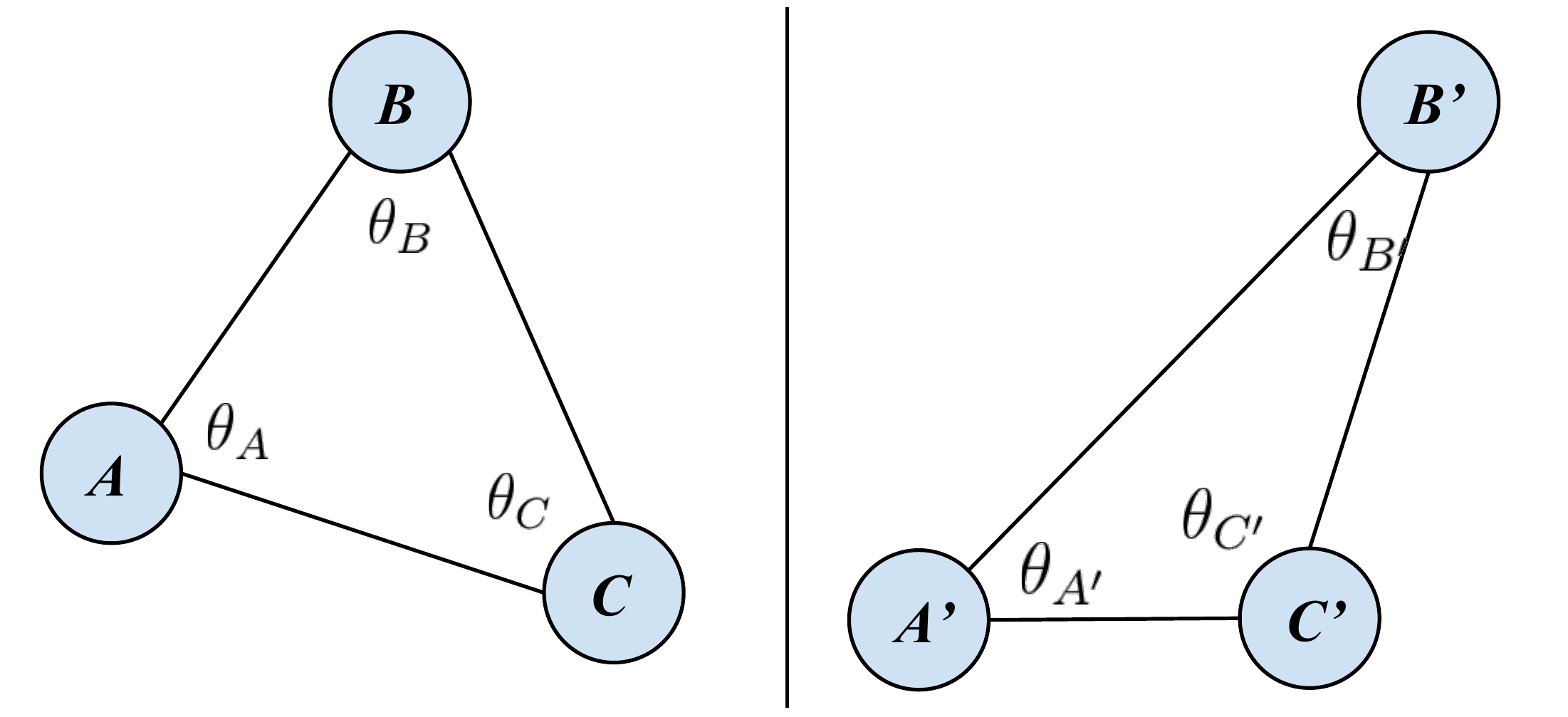}
\caption{\label{fig3} Visualization of the shape constraint. The constraint compares the three angles of two
triplets.}
\end{figure}

\noindent {\bf The shape constraint} measures the shape of the triangle defined by three patch centers (Figure \ref{fig3}). Let
$\Theta_{ABC} = \{\theta_A, \theta_B,\theta_C\}$ denote the angles of triangle $ABC$, and 
$\Theta_{A'B'C'} = \{\theta_{A'}, \theta_{B'}, \theta_{C'}\}$ denote the angles of triangle $A'B'C'$, as displayed
in Figure \ref{fig3}. We define a shape penalty function by comparing corresponding angles as
\begin{align}
\nonumber
& p_s(\Theta_{A'B'C'}, \Theta_{ABC})  \\
 = & 1 - \eta_s \frac{\sum_{i \in \{A, B, C\}} \lvert\cos(\theta_i) - \cos(\theta_{i'}) \rvert} {6}, 
\label{eq4}
\end{align}
where $\eta_s \in [0, 1]$ controls how important the shape constraint is. The denominator $6$ in Eq.~(\ref{eq4}) ensures
that $0 \le p_s \le 1$, since $\lvert\cos(\theta_i) - \cos(\theta_{i'}) \rvert \le 2$. $\{\cos(\theta_i)\}$ and
$\{\cos(\theta_{i'})\}$ can be easily computed from inner products. The
motivation for introducing this second constraint is that we use the relatively loose order constraint to perform coarse 
verification and use the shape constraint for finer adjustments. As will be demonstrated in Section \ref{sec4_1}, the
two constraints contain complementary information.

\subsection{Triplet Detector} \label{sec2_2}
Our triplet detector consists of three appearance models and the two geometric constraints. We
construct three linear weights $\{w_A, w_B, w_C\}$ from the patch templates $\{T_A, T_B, T_C\}$, and our triplet
detector becomes 
\begin{equation}\label{eq+1}
{\bf T} = \{w_A, w_B, w_C, G_{ABC}, \Theta_{ABC}\}. 
\end{equation}
For any triplet $\{A', B', C'\}$ with features $\{T_{A'}, T_{B'}, T_{C'}\}$ and geometric parameters $\{G_{A'B'C'},
\Theta_{A'B'C'}\}$, its detection score is defined as
\begin{align}
S_{A'B'C'} 
= (S_A(T_A') + S_B(T_B') + S_C(T_C'))\cdot p_o \cdot p_s,\label{eq5}
\end{align}
where $p_o$, $p_s$ are defined by Eq.~(\ref{eq2}), Eq.~(\ref{eq4}) respectively, and the appearance scores are defined
as
\begin{equation} \label{eq6}
S_i(T_{i'}) = w_i^{\rm T}T_{i'}, \quad i \in \{A, B, C\}.
\end{equation}

To make triplet detection practical, three technical details must be addressed. The first is how to efficiently obtain the maximum
response of a triplet detector in an image. In principle, we could examine all possible triplets from the combinations
of all possible patches and simply compute 
\begin{equation}\label{eq7}
\{A^{*}, B^{*}, C^{*}\} = \argmax_{\{A', B', C'\}} S_{A'B'C'}.
\end{equation}
However, this is too expensive since the number of all possible triplets will be $O(N^3)$ for $N$ patches. Instead, we
adopt a greedy approach. We first find the top $K$ non-overlapping
detections for each appearance detector independently. Then we evaluate the $K^3$ possible triplets and
select the one with the maximum score defined by Eq.~(\ref{eq5}).

The second technical detail is how to obtain the linear weights $w_i$ from a patch template $T_i$. For efficiency, we
use the LDA model introduced by \cite{ldadet}
\begin{equation} \label{eq8}
w_i = \Sigma^{-1} \left(T_i - \mu \right),
\end{equation}
where $\mu$ is the mean of features from all patches in the dataset, and $\Sigma$ is the corresponding covariance matrix.
The LDA model is efficient since it constructs the model for negative patches ($\mu$ and $\Sigma$) only once.

Our triplet detector is able to handle moderate pose variations. However, if we flip an image, both the appearance (\eg
the dominant direction of an edge) and the order of the patches will change. We address this issue by applying the
triplet detector to the image and its mirror, generating two detection scores, and choose the larger score as the
response. This simple technique proves effective in practice.

\section{Discriminative Triplets for Fine-Grained Classification} \label{sec3}
In this section, we describe how to automatically mine discriminative triplets with the geometric constraints and
generate mid-level representations for classification with the mined triplets. We present the overview of our framework
in Figure~\ref{fig4}. In the triplet
initialization stage, we use a nearest-neighbor approach to propose
potential triplets, taking advantage of the fact that instances of fine-grained objects share similar overall appearance. Then we verify the
discriminativeness of the candidate triplets using the whole training set or a large portion of it, and select
discriminative ones according to an entropy-based measure. For classification, we concatenate the maximum responses of the selected
discriminative triplets to construct mid-level image representations. The key to our approach
is proposing candidate triplets locally and selecting discriminative ones globally, avoiding the insufficient
data problems of other nearest-neighbor based fine-grained approaches.

\subsection{Triplet Initialization} \label{sec3_1}
To reduce the computational burden of triplet mining, we initialize candidate triplets using potentially discriminative
patches in a nearest-neighbor fashion. The overview of the procedure is displayed in Figure~\ref{fig6}.

\begin{figure}
\centering
\includegraphics[width=0.48\textwidth]{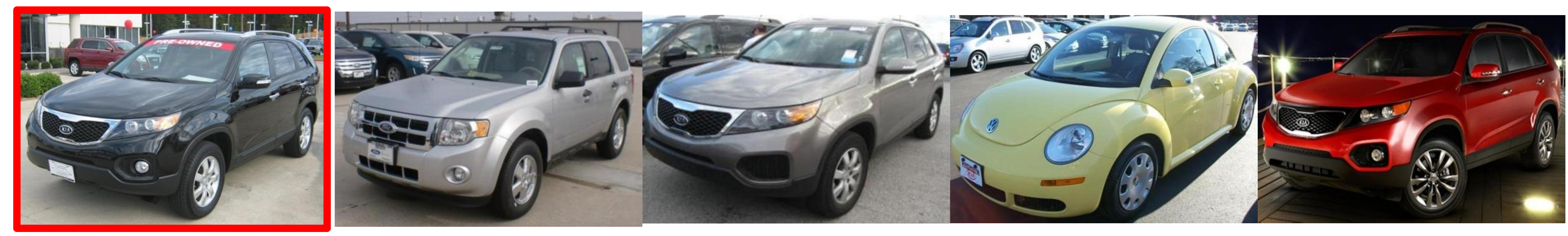}
\caption{\label{fig5} Some examples of a set of neighboring images. The query image is highlighted with
a red box. Since fine-grained objects share similar overall appearance, the neighborhood consists of samples from
different classes with the same pose.}
\end{figure}

\begin{figure*}
\centering
\includegraphics[width=0.92\textwidth]{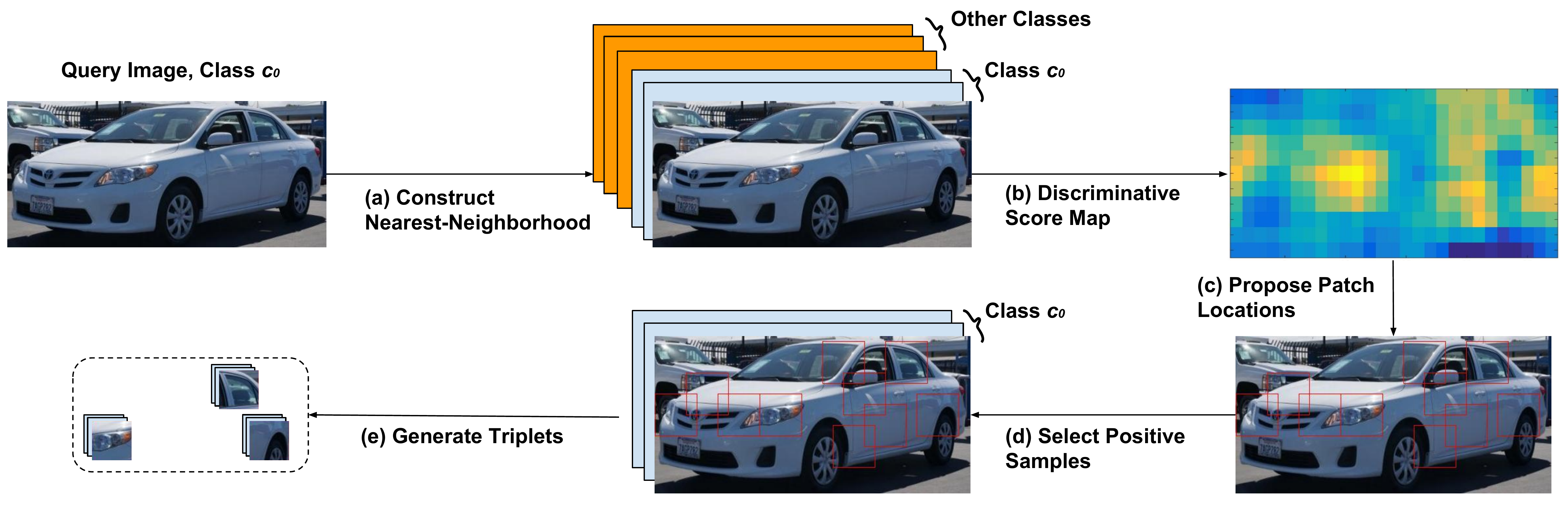}
\caption{\label{fig6} Visualization of the triplet initialization stage. {\bf(a)} A seed image from class $c_0$ is used to
construct its nearest-neighbor set including itself. {\bf(b)} Discriminative score map is generated from the stack of neighboring images.
{\bf(c)} Patch locations with top discriminative scores are selected by non-maximum suppression. {\bf(d)} Images from positive class
(class $c_0$) are selected to generate triplets. {\bf(e)} Triplets are generated from positive samples at locations proposed
in (c).}
\end{figure*}

\vspace{0.2em}
\noindent{\bf Construct Neighborhood.} \quad For a seed training image $I_0$ with class label $c_0$, we extract features
\cite{hog} of the
whole image $X_0$ and use it to retrieve the nearest neighbors from the training set. Since fine-grained objects have
similar overall appearance, the resulting set of images consists of training images from different classes with almost
the same pose (Figure~\ref{fig5}). This first step results in a set of roughly aligned images, so that potentially
discriminative regions can be found by comparing corresponding regions across the images. We refer to the set consisting
of a seed training image and its nearest neighbors as a neighborhood.

\vspace{0.2em}
\noindent{\bf Find Candidate Regions.} \quad We regard each neighborhood as a stack of aligned images with their class
labels and locate potentially
discriminative regions. Consider the set of patches at the same location of each image. For each location $(x, y)$, let
$F_i(x, y)$ be the features extracted from the patch in the $i^{th}$ image and
let $c_i$ be its label. Denote the set of observed class labels as $C$. The discriminative score at $(x, y)$ is simply defined as
the ratio of between-class variation and in-class variation, {\it i.e.} 
\begin{equation} \label{eq9}
d(x, y) = \frac{\sum_{c \in C} \left\| \overline{F_c}(x, y) - \overline{F}(x, y)\right\|^2} {\sum_{c \in C} \sum_{c_i = c} \left\|
F_i(x, y) -
\overline{F_c}(x, y)\right\|^2},
\end{equation}
where $\overline{F}(x, y)$ is the averaged feature of all patches at location $(x, y)$, and $\overline{F_c}(x, y)$ is
the average of patches from class $c$.  
We compute discriminative scores $d(x, y)$ for patch locations in a sliding window fashion, with patch size $64\times 64$ and
stride $8$, and choose the patch locations with top scores. To ensure the diversity of the
regions, non-maximum suppression is used to allow only a small amount of overlap. 

\vspace{0.2em}
\noindent{\bf Propose Candidate Triplets.} \quad For a neighborhood generated from the seed image with class label $c_0$, candidate triplets are
proposed as follows. We first select all positive samples with label $c_0$ in the neighborhood. 
For each positive sample, we extract features from patches at the discriminative patch locations obtained in
the last step. Then for patch location $i$, the patch template $T_i$ is obtained by averaging the features of all the positive samples. We
propose candidate triplets by selecting all possible combinations of three patch locations. To avoid duplicate triplets,
we rank three patch locations by their discriminative scores Eq.~(\ref{eq9}). We
construct triplet detectors with geometric constraints from these patch triplets as discussed in Section \ref{sec2_2}. 

In practice, in each neighborhood we find the top $6$ discriminative locations and propose $\binom{6}{3} = 20$ candidate triplets. By
considering all the neighborhoods for every class, we obtain the pool of candidate triplets. 

\subsection{Discriminative Triplets Mining by Entropy Scores} \label{sec3_2}
Candidate triplets are constructed from potentially discriminative regions measured by Eq.~(\ref{eq9}). However, this
measure is computed only within a small neighborhood and might be noisy due to lack of data. On the other hand,
recent approaches to scene understanding \cite{patch1, patch2, patch4, patch3} mine discriminative patches using a large portion of
the training data and obtain very good results. Consequently, we select discriminative triplets by evaluating each
candidate triplet on the broader dataset.

For each triplet detector obtained in Section \ref{sec3_1}, we detect triplets in each training image and
obtain the maximum detection score as discussed in Section \ref{sec2_2}, {\it i.e.}, find the top $K$ detections for
each appearance detector, consider all $K^3$ triplets, and choose the one with maximum geometrically-penalized score
Eq.~(\ref{eq5}).
We obtain the top detections within the training set along with their corresponding class labels. If a triplet is 
discriminative across the training set, the top detections are expected to arise from only one or a few classes. Therefore, if
we calculate the entropy of the class distribution over the top detections, the entropy should be low.
Let $p(c|{\bf T})$ denote the probability of top detections coming from class $c$ for triplet $\bf T$. Then
\begin{equation} \label{eq10}
H(c|{\bf T}) = \sum_{c} p(c|{\bf T}) \log p(c|{\bf T})
\end{equation}
is an entropy-based measure that has been effectively used by \cite{patch3, flh} for patch selection. We
calculate this measure for all candidate triplets and choose the ones with the lowest entropy
to form the set of discriminative triplets.
\begin{figure*}
\centering
\centering
\includegraphics[width=\textwidth]{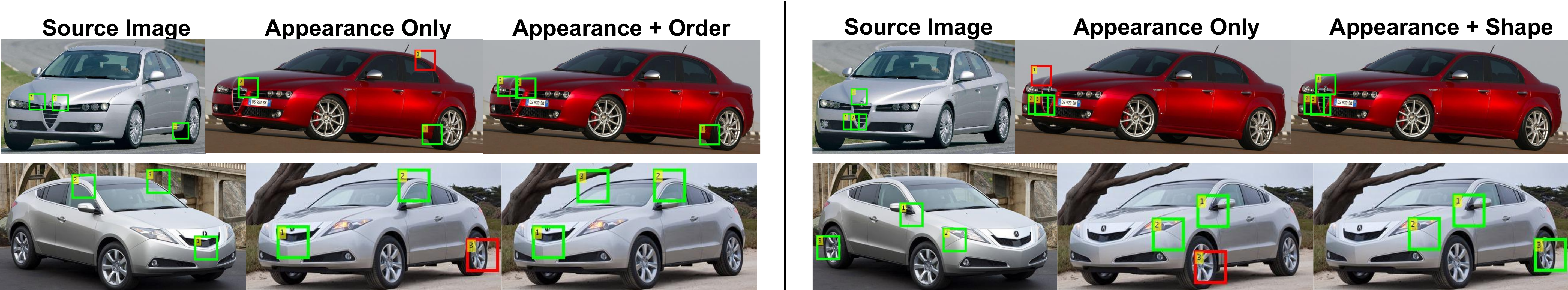}
\caption{\label{fig7} Visualization of the effects of the two geometric constraints. Red boxes are incorrectly localized
patches. The order constraint roughly checks
the geometric arrangement of three patches and can eliminate incidental false detections which happen to have high
appearance score; the shape constraints enforce finer adjustments on patch locations than the order constraint.}
\end{figure*}
\subsection{Mid-Level Image Representations for Classification}
Finally, we use the maximum responses of the mined discriminative triplet detectors to construct a mid-level representation for an image.
The dimension of the mid-level representation equals the number of triplet detectors, with each detection score
occupying one dimension in the image-level descriptor. The mid-level representation is used as the input for a linear
SVM to produce the classification result. We refer to the image-level descriptor as a {\it Bag of Triplets} (BoT).

\section{Experiments} \label{sec4}
\subsection{Triplet Localization with Geometric Constraints} \label{sec4_1}
We first design a simple experiment to demonstrate that the geometric constraints described in Section
\ref{sec2} can actually improve patch localization, assuming we already have a good patch set. To achieve this
goal, we use the FG3DCar dataset provided by \cite{yenliang}. This dataset consists of 300
car images from 30 classes, with each image annotated with 64 ground-truth landmark points. Cars in this
dataset have large pose variations, which makes triplet localization difficult. 

\vspace{0.2em}
\noindent {\bf Experimental Settings.} \quad The experiment is designed as follows. For each image, we construct a good set of 64
patches by extracting the ones located at the annotated landmarks. We repeatedly and randomly select two images from the
same class and obtain two corresponding sets of 64 patches with their locations. Then we randomly select three patches
from one image (denoted as Image 1) to construct a triplet detector in Eq.~(\ref{eq+1}) and attempt to find the corresponding
triplet in the pool of 64 patches of the other (denoted as Image 2). During the experiment, patches are represented
by the Fisher Vector features provided by \cite{yenliang}.

The following four methods are evaluated on this task. The decision procedure of the three methods with gemetric
constraints is discussed in Section \ref{sec2_2}.
\begin{itemize}
\item Appearance Only (Baseline): Independently apply each patch detector, and choose the detection with
the highest score. The detected triplet consists of the three top individual detections. In this method, only the
appearance features of the three patches are used.
\item Order Constraint: Use the appearance and the order constraint by
setting $\eta_s = 0$ in Eq.~(\ref{eq4}), such that $p_s = 1$ (no shape penalty) always holds in Eq.~(\ref{eq5}). 
\item Shape Constraint: Use the appearance and the shape constraint by
setting $\eta_o = 0$ in Eq.~(\ref{eq2}), such that $p_o = 1$ (no order penalty) always holds in Eq.~(\ref{eq5}). 
\item Combined: Use both geometric constraints. In practice, we set $\eta_o = 0.5$ and $\eta_s = 1$.
\end{itemize}

Due to large pose changes, several landmark locations are highly overlapped with each other in some images.
Therefore, during evaluation, each patch is regarded as correctly localized if the detected patch is (i) the same as
the ground
truth corresponding patch; (ii) highly overlapped with the ground truth corresponding patch, with overlap/union
ratio greater than 50\%. Each triplet is regarded as successfully localized if all of its three patches are correctly
localized. We randomly select 1000 image pairs, and for each pair we randomly test 1000 triplets. The accuracy of
triplet localization is the percentage of successfully localized triplets over all the 1 million triplets evaluated.
\begin{table}
\centering
\begin{tabular}{|c|c|c|}
\hline
Method & \parbox[t]{2cm}{Localization \\Accuracy (\%)} & \parbox[t]{2.6cm}{Improvement \\Over Baseline (\%)}\\
\hline \hline 
Appearance Only & 24.9 & - \\
\hline 
Order Constraint & 27.7 & 11.2 \\
\hline
Shape Constraint & 34.4 & 38.2 \\
\hline
Combined & 35.3 & 41.9 \\
\hline
\end{tabular}
\vspace{0.5em}
\caption{\label{tb1} Triplets localization test result on FG3DCar dataset. The localization accuracy and relative
improvement over baseline (Appearance Only method) are demonstrated.}
\end{table}

\vspace{0.2em}
\noindent {\bf Result and Analysis.} \quad We demonstrate triplet localization accuracy
and relative improvement over baseline in Table \ref{tb1}. Even though we have a human-annotated pool of patches, localization is
challenging with appearance only, since the pose variations are large and the appearance detector is learnt
with only one positive sample. As we add geometric constraints, we obtain cumulative improvement over
the baseline. Typical examples indicating the
effects of the two constraints are displayed in Figure \ref{fig7}. The order constraint, which is
relatively loose, tends to roughly check the geometric arrangement of three patches. It can eliminate the patches which
happen to have a very high appearance score. On the other hand, the shape constraint enforces fine adjustment, which is
complementary to the order constraint. With the two geometric constraints combined, the improvement
is significant.

\subsection{Fine-Grained Classification}\label{sec4_2}
We demonstrate fine-grained classification
results on three standard fine-grained car datasets. No extra annotation beyond object bounding boxes is used throughout the experiments.
When comparing results, we refer to our approach as {\it Bag of Triplets} (BoT).

\subsubsection{14-Class BMVC Cars Dataset Results}\label{sec4_2_1}

{\bf Dataset.} \quad The fine-grained car dataset provided by \cite{bmvccar} (denoted as BMVC-14) consists
of 1904 images of cars from 14 classes. \cite{bmvccar} has split the data into 50\% {\it train}, 25\% {\it
val} and 25\% {\it test}. We follow this setting for evaluation.

\vspace{0.2em}
\noindent {\bf Experimental Settings.} \quad The implementation details of our discriminative triplet mining approach
are briefly stated as follows. Each image is cropped to its bounding box and resized such that the width is 500 (aspect ratio
maintained). The patch size is set to be $64 \times 64$ and HOG features are extracted to represent the
patches for fair comparison to preivously reported results. In the triplet initialization stage, for each seed image we
construct the neighborhood of size
20 including itself. As mentioned in Section \ref{sec3_1}, for each neighborhood we propose the top 6 discriminative patch locations and
propose $\binom{6}{3} = 20$ triplets for mining. In the triplet mining stage, we obtain the
top detections across the whole training set and calculate entropy measure Eq.~(\ref{eq10}). Then we select 300
discriminative triplets per class. Finally, the mid-level representation has dimension $14\times 300 = 4200$, which is
fed into the linear SVM implemented by LIBLINEAR \cite{liblinear}. 

We test the following two cases:
\begin{itemize}
\item Without Geometric Constraints (Without Geo): In the discriminative mining and mid-level representation
construction stages, we adopt the ``Appearance Only'' triplet detection strategy described in Section \ref{sec4_1}. 
\item With Geometric Constraints (With Geo): Each time
we use a triplet detector, we use Eq.~(\ref{eq5}) and related techniques in Section \ref{sec2_2} to incorporate the two constraints. By
comparing the two cases, we quantitatively test the effectiveness of geometric constraints.
\end{itemize}

\noindent {\bf Results and Analysis.} \quad We compare our results with previous work, citing the results from
\cite{yenliang}, which has provided a summary of previously published results on BMVC-14. It includes several baseline methods such as LLC
\cite{llc} and PHOW \cite{phow} with codebook size 2048, Fisher Vector (FV) \cite{fishervec} with 256 Gaussian Mixture Model
(GMM) components, as well as structDPM \cite{bmvccar} and BB-3D-G \cite{car196} specifically designed for the task.
Among these methods, BB-3D-G \cite{car196} used extra 3D models, while others only used ground
truth bounding boxes as we did. The results are summarized in Table \ref{tb2}. Our method without geometric
constraints outperforms all three baseline methods. When geometric constraints are further added, our
approach not only outperforms the best reported result using only bounding boxes with a noticeable margin, but it also
outperforms the BB-3D-G method which uses extra 3D model fitting. It is worth mentioning that our method with triplet
geometric constraints outperforms the DPM-based method \cite{bmvccar}, which incorporates root-part pair-wise constraints.

\begin{table}
\centering
\begin{tabular}{|p{4.5cm}|c|}
\hline
Method & Accuracy (\%)\\
\hline
\hline
LLC \cite{llc} & 84.5 \\
PHOW \cite{phow} & 89.0\\
FV \cite{fishervec} & 93.9\\
structDPM \cite{bmvccar} & 93.5\\
BB-3D-G \cite{car196} & 94.5\\
\hline
BoT (HOG Without Geo) & 94.1\\
BoT (HOG With Geo) & 96.6\\
\hline
\end{tabular}
\vspace{1em}
\caption{\label{tb2} Results on BMVC-14 dataset.}
\end{table}
\begin{figure}
\centering
\includegraphics[width=0.32\textwidth]{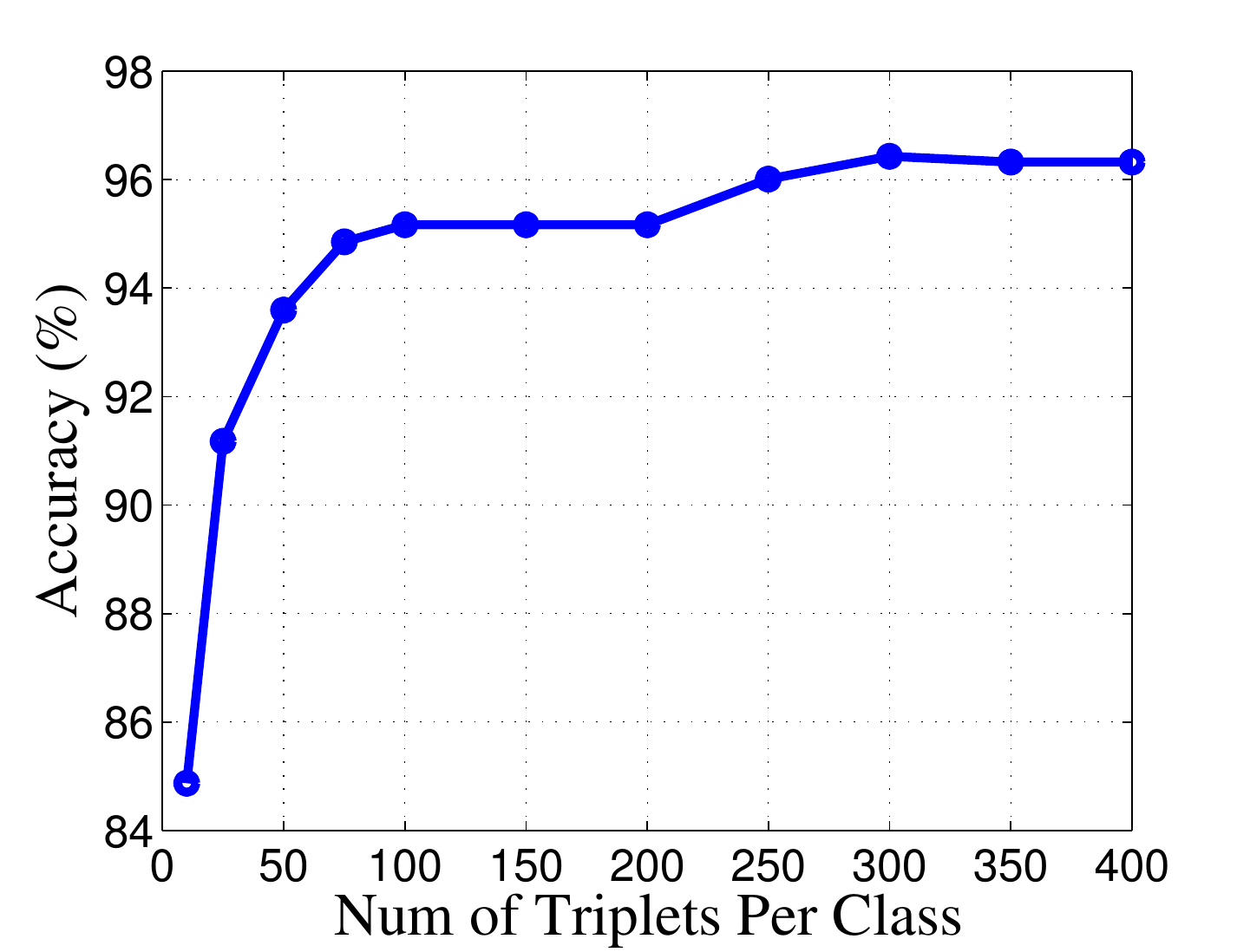}
\centering
\caption{\label{fig8} Classification accuracy with respect to the number of triplets per class.}
\end{figure}
\begin{figure*}
\centering
\includegraphics[width=\textwidth]{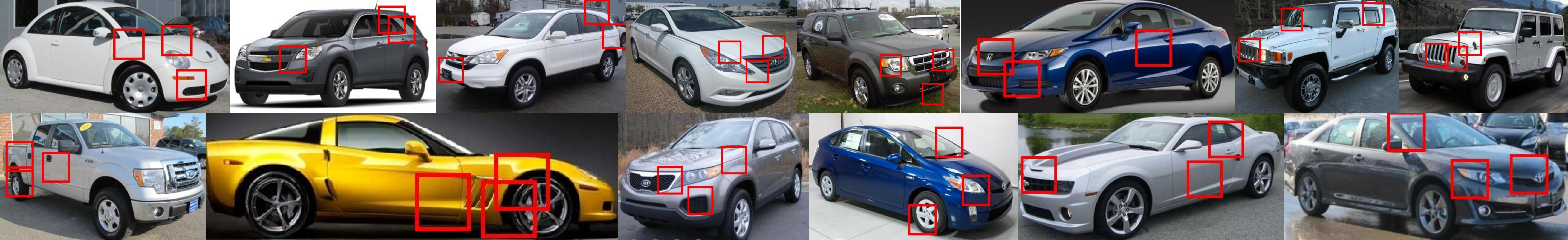}
\caption{\label{fig9}Visualization of the most discriminative triplet (measured by Eq.~(\ref{eq10})) for each class in
BMVC-14 proposed by our method. The triplets accurately capture the subtle discriminative information of each class,
which is highly consistent with human perception. For instance, for the \textbf{first image in the first row}, the triplet captures the 
curvy nature of Volkswagen Beetle such as rounded hood; for the \textbf{first image in the second row}, the triplet focuses on
the rear cargo of the pick-up truck, since Ford F-Series is the only pick-up in the dataset; for the \textbf{last image in the
first row}, the triplet highlights the frontal face of Jeep Wrangler.}
\end{figure*}

We plot the performance as a function of the number of
discriminative triplets per class (with geometric constraints) in Figure \ref{fig8}. When we
use only 10 triplets/class, the performance of $84.9 \%$ already outperforms the baseline LLC \cite{llc}, suggesting that the
mined discriminative triplets are highly informative. As we increase the number of triplets per class, performance more
or less saturates after 100 triplets/class, although the best performance is at 300 triplets/class. Therefore, when we 
deal with large-scale datasets such as the Stanford Cars dataset below, we can use a smaller number of triplets
to construct lower-dimensional mid-level descriptors without much loss in performance. As the number of triplets/class
exceeds 300, the performance decreases, suggesting that the remaining triplets, which rank low by our criteria, do not
add discriminatively useful information.

We further visualize the most discriminative triplet measured by the entropy score Eq.~(\ref{eq10}) for all 14 classes in
Figure~\ref{fig9}. The triplets in the figure accurately localize the subtle discriminative regions, which are highly
consistent with human perception, such as the distinctive side vent grill of Chevrolet Corvette (the second image in the
second row of Figure~\ref{fig9}, see Figure~\ref{fig9} for more details). This empirically explains why our triplets are highly informative.
\subsubsection{196-Class Stanford Cars Dataset Results}\label{sec4_2_2}

{\bf Dataset.} \quad The Stanford Cars Dataset \cite{car196} contains 16,185 car images from 196 classes (denoted
as Cars-196). The data
split provided by \cite{car196} is 8,144 training images and 8,041 testing images, where each class has been
roughly split 50-50. We follow this setting in our experiment.

\vspace{0.2em}
\noindent {\bf Experimental Settings.} \quad Our method focuses on generating effective mid-level representations and is
independent from the choice of low-level features. In this experiment, in order to fairly compare to both the
traditional methods without using extra data/annotation and the more recent ones which finetune ImageNet pre-trained
Convolutional Neural Networks (CNN), we evaluate our approach using both HOG and
CNN features as the low-level representations of the patches. When extracting CNN features, we directly use the off-the-shelf
ImageNet pre-trained CNN model as a general feature extractor without any finetuning. For fair comparison, we
adopt the popular 16-layer VGGNet-16 \cite{vgg} as the network architecture, and extract features from $\text{pool}_4$
layer, which is the max-pooled output of its $10^{\text{th}}$ convolutional layer.

Also, we adapt our approach slightly to handle such a large-scale dataset. Instead of traversing the whole dataset, when
retrieving nearest-neighbors for a seed image with class label $c_0$, we regard class $c_0$ as the positive class,
randomly select 29 other classes as negative classes, and retrieve nearest-neighbors within the training images from
these 30 classes; when finding top detections for a triplet from class $c_0$, we use the training
images from class $c_0$ and 14 randomly selected negative classes.
Additionally, since we have empirically determined that the discriminative triplets are informative in Figure \ref{fig8}, we select
150 discriminative triplets per class. 

Except for the settings described above, other parameters remain the same as those in Section \ref{sec4_2_1}.
\begin{figure*}
\centering
\includegraphics[width=0.85\textwidth]{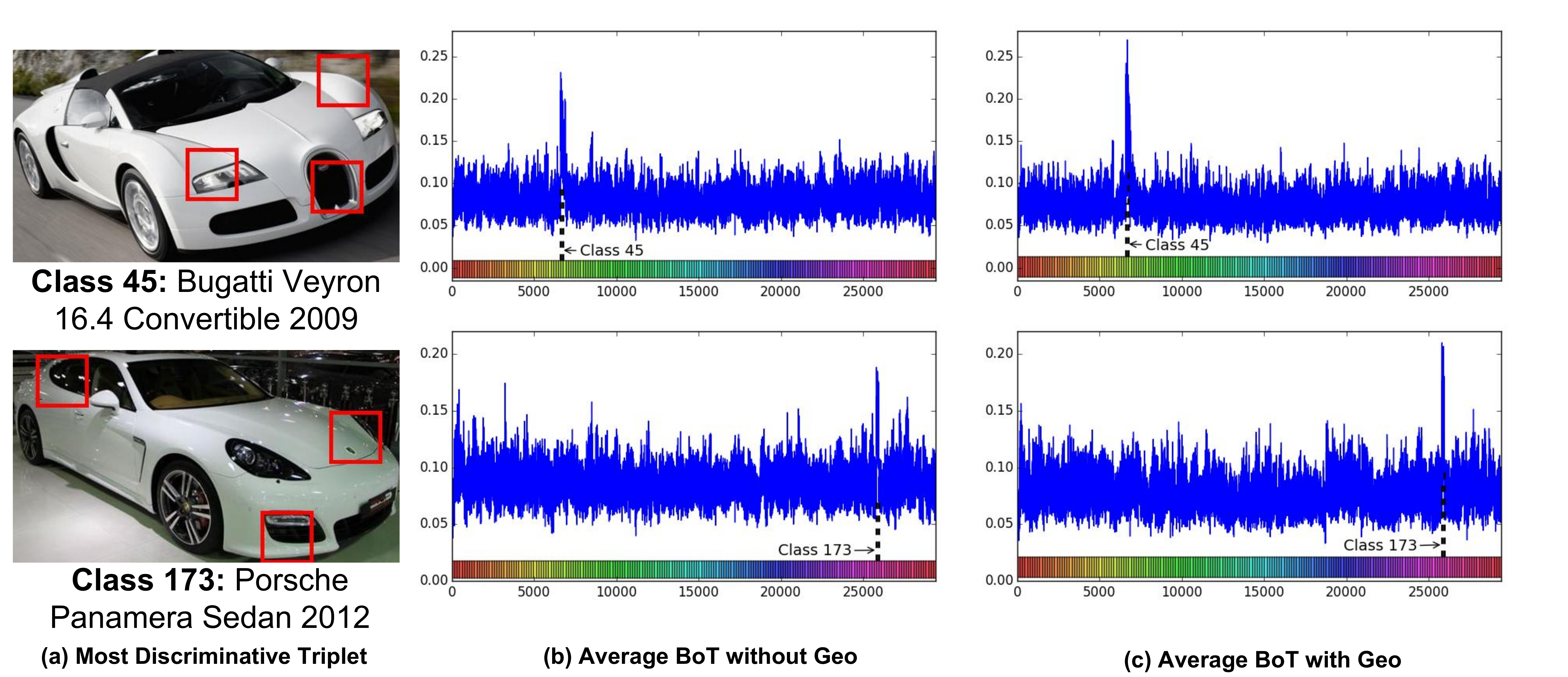}
\caption{\label{fig10} (a) Visualization of the most discriminative triplets of two example classes in Cars-196. (b)(c) The averaged
image-level BoT descriptor across all test samples in the corresponding class. Each dimension in the BoT is generated by the
response of a mined triplet. The color bars are used to describe the dimensions correponding to the responses of triplets from different classes.}
\end{figure*}

\noindent {\bf Results and Analysis.} \quad Our baselines
include LLC \cite{llc} as HOG-based baseline and AlexNet \cite{alexnet} as CNN baselines. For CNN, we cite the
result of training an AlexNet from scratch on Cars-196 without extra data (AlexNet From
Scratch) \cite{ellf} and the result of finetuning an ImageNet pre-trained AlexNet on Cars-196 (AlexNet Finetuned) \cite{har}.
We also compare with previously published results including BB-3D-G \cite{car196},
ELLF \cite{ellf}, FT-HAR-CNN \cite{har}, Bilinear CNN (B-CNN) \cite{b_cnn} and the method with the highest reported
accuracy so far \cite{krause15}. It is worth mentioning that the last three approaches \cite{har, b_cnn, krause15} are
CNN-based, where \cite{har} is AlexNet based, \cite{b_cnn} is VGGNet-16 based, and \cite{krause15} is 19-layer VGGNet-19
based. \cite{har} finetunes a CNN
with the help of another 10,000 images of cars without fine-grained labels; the best result of \cite{b_cnn} is achieved
by finetuning a two-stream CNN architecture; \cite{krause15} integrates segmentation, graph-based alignment, finetuned
R-CNN \cite{rcnn} and SVM to produce its best result.

The results are displayed in Table \ref{tb4}. Even though we adopted relatively ``economical''
settings, our method behaves stably and operates at the state-of-the-art performance. When using HOG as low-level patch
representation, our approach not only greatly outperforms the HOG-based baseline (LLC) (by more than 15\%), but it 
even outperforms the CNN baseline of finetuned AlexNet by a fairly noticeable margin (more than 2\%) -- a significant achievement since we are
only using HOG and geometric constraints without any extra data or annotations. When using off-the-shelf CNN features,
our method with geometric constraints outperforms B-CNN \cite{b_cnn} which uses two streams of VGGNet-16, and obtained
quite comparable results to the state-of-the-art \cite{krause15}. Furthermore, our method does not perform finetuning and depends on the
strength of our discriminative triplet mining itself, which is supported by the results, rather than the learning capability of CNNs.  

To intuitively demonstrate the effectiveness of the geometric constraints, we plot the image-level BoT descriptors from a few
classes in the second and third columns of Figure \ref{fig10}. For each class we plot the averaged BoT descriptor across
all {\it test} samples from that class. 
Figure \ref{fig10} shows that after introducing the geometric constraints, the BoT descriptor becomes more
peaked at the corresponding class, since the geometric constraints help learn more discriminative triplets which
generate more peaky responses, as well as penalizing those incorrect detections from other classes which happen to have high
appearance scores (which can be clearly seen from the second row of Figure \ref{fig10}). This 
discriminative capability is achieved during test time, showing that our approach generalizes very well.

Finally, we visualize the most discriminative triplet measured by Eq.~(\ref{eq10}) in the first column of Figure
\ref{fig10}. Similar to Figure \ref{fig9}, our approach captures the subtle difference of fine-grained
categories and accurately localizes the discriminative regions, which are highly interpretable by humans.
For example, it highlights the distinctive air grill and rounded fender of Bugatti Veyron, and the classical headlight
and tail of Porsche.

\begin{table}
\centering
\begin{tabular}{|p{4.5cm}|c|}
\hline
Method & Accuracy (\%) \\
\hline\hline
LLC$^*$\cite{llc} & 69.5 \\
BB-3D-G \cite{car196} & 67.6 \\
ELLF$^*$ \cite{ellf} & 73.9 \\
AlexNet From Scratch \cite{ellf} & 70.5 \\
AlexNet Finetuned \cite{har} & 83.1\\
FT-HAR-CNN \cite{har} & 86.3 \\
B-CNN \cite{b_cnn} & 91.3 \\
Best Result in \cite{krause15} & 92.8 \\
\hline
BoT(HOG Without Geo)$^*$ & 84.6 \\
BoT(HOG With Geo)$^*$ & 85.7 \\
BoT(CNN Without Geo) & 91.2\\
BoT(CNN With Geo) & 92.5\\
\hline
\end{tabular}
\vspace{1em}
\caption{\label{tb4} Results on Cars-196 dataset. Items with ``*'' indicate that no extra annotations/data are involved.}
\end{table}

\subsubsection{100-Class FGVC-Aircraft Dataset Results}\label{sec4_2_3}
Finally, to demonstrate that our approach is effective in multiple fine-grained domains, we briefly present our results on
FGVC-Aircraft dataset
\cite{fgvc_air}, which contains 10,000 images from 100 classes of aircrafts and is of similar scale to
the Cars-196 dataset (16185 images from 196 classes). For fair comparison, we use the standard train/test split provided by
the dataset provider \cite{fgvc_air} and the parameter settings of our approach are \textit{exactly the same} as those in Section
\ref{sec4_2_2}. We report our results in Table \ref{tb5}. Our approach using CNN features (without
fine-tuning) outperforms state-of-the-art (VGGNet-16 based) \cite{b_cnn} by a noticeable margin. The results
suggest that our approach performs well in various fine-grained domains.
\begin{table}
\centering
\begin{tabular}{|p{4.5cm}|c|}
\hline
Method & Accuracy (\%) \\
\hline
\hline 
Symbiotic \cite{symbiotic} & 75.9 \\
Fine-tuned AlexNet \cite{fgvcfisher} & 78.9 \\
Fisher Vector \cite{fgvcfisher} & 81.5 \\
B-CNN \cite{b_cnn} & 84.1 \\
\hline 
BoT (CNN without Geo) & 86.7 \\
BoT (CNN with Geo) & 88.4 \\
\hline 
\end{tabular}
\vspace{0.5em}
\caption{\label{tb5}Results on FGVC-Aircraft dataset.}
\end{table}

\section{Conclusion} \label{sec5}
We proposed a mid-level patch-based approach for fine-grained classification. We first introduce triplets of patches
with two geometric constraints to improve localizing patches, and automatically mine discriminative triplets to
construct mid-level representations for fine-grained classification. Experimental results demonstrated
that our discriminative triplets mining framework performs very well on both mid-scale and large-scale fine-grained
classification datasets, and outperforms or obtains comparable results to the state-of-the-art. 

\noindent {\bf Acknowledgements}\quad This work was partially supported by ONR MURI Grant N000141010934.

{\small
\bibliographystyle{ieee}
\bibliography{egbib}
}

\end{document}